\documentclass[conference]{IEEEtran}
\usepackage{times}

\usepackage[numbers]{natbib}
\usepackage{multicol}
\usepackage[bookmarks=true]{hyperref}
\usepackage{array}
\usepackage{multirow}
\usepackage{makecell} 
\usepackage{colortbl}
\usepackage{xcolor}
\usepackage{xspace}

\usepackage{booktabs}
\usepackage{siunitx}   
\usepackage{booktabs}
\usepackage[table]{xcolor} 

\usepackage{tabularx}   
\usepackage{array}      
\usepackage{siunitx}

\usepackage{multirow}
\usepackage{bbding}
\usepackage{graphicx}
\usepackage{float}
\usepackage{cuted}
\usepackage{caption}
\usepackage[most]{tcolorbox}

\usepackage{amsmath}

\newcommand{\ours}{CaPE}

\begin{document}

\title{Safe and Interpretable Multimodal Path Planning for Multi-Agent Cooperation}

\author{
\IEEEauthorblockN{
Haojun Shi\thanks{Equal contribution.}$^{*}$,
Suyu Ye$^{*}$,
Katherine M. Guerrerio,
Jianzhi Shen,
Yifan Yin,
Daniel Khashabi,\\
Chien-Ming Huang,
Tianmin Shu
}
\IEEEauthorblockA{
Johns Hopkins University
}
}

\twocolumn[{%
\renewcommand\twocolumn[1][]{#1}%
\maketitle
\vspace{-10pt}
\begin{center}
    \centering
    \captionsetup{type=figure}
     \includegraphics[trim=0cm 14.8cm 0cm 3.5cm, clip, width=\textwidth]{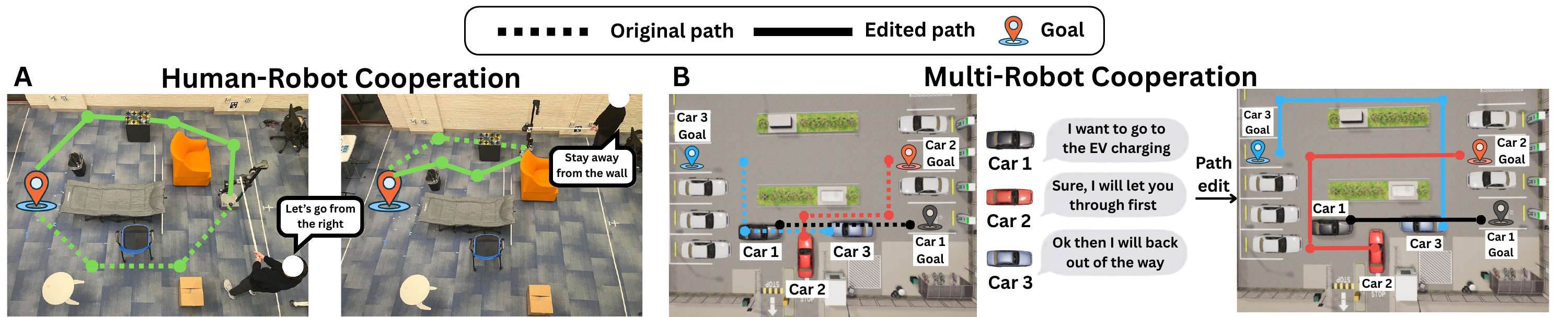}
\captionof{figure}{Examples of grounding natural language communication into path edits for different multi-agent cooperation scenarios. (A) Human-robot cooperation in a joint-lifting setting. The robot sequentially edits its path in response to human instructions. (B) Multi-robot coordination. After language-based negotiation, the robots update their paths accordingly to avoid collisions. Dashed lines indicate original paths, and solid lines indicate the edited paths after grounding the communication.
}
\label{fig:concept}
\end{center}
\vspace{0.05in}
}]

\renewcommand{\thefootnote}{}
\footnotetext{$^*$Equal Contribution.}
\footnotetext{Code and data are available at \url{https://scai.cs.jhu.edu/projects/CaPE}.}
\renewcommand{\thefootnote}{\arabic{footnote}}

\begin{abstract}
Successful cooperation among decentralized agents requires each agent to quickly adapt its plan to the behavior of other agents. In scenarios where agents cannot confidently predict one another's intentions and plans, language communication can be crucial for ensuring safety. In this work, we focus on path-level cooperation in which agents must adapt their paths to one another in order to avoid collisions or perform physical collaboration such as joint carrying. In particular, we propose a safe and interpretable multimodal path planning method, \ours~(Code as Path Editor), which generates and updates path plans for an agent based on the environment and language communication from other agents. \ours~leverages a vision-language model (VLM) to synthesize a path editing program verified by a model-based planner, grounding communication to path plan updates in a safe and interpretable way. We evaluate our approach in diverse simulated and real-world scenarios, including multi-robot and human-robot cooperation in autonomous driving, household, and joint carrying tasks. Experimental results demonstrate that \ours~can be integrated into different robotic systems as a plug-and-play module, greatly enhancing a robot's ability to align its plan to language communication from other robots or humans. We also show that the combination of the VLM-based path editing program synthesis and model-based planning safety enables robots to achieve open-ended cooperation while maintaining safety and interpretability.

\end{abstract}

\IEEEpeerreviewmaketitle

\section{Introduction}

Robot path planning for multi-agent cooperation is a common challenge between a human and a robot or between multiple robots \cite{dahiya2023survey, sharon2015conflict, wang2025paths}. It can encompass a wide range of multi-agent scenarios. In some cases, agents with heterogeneous goals that share a common space must adjust their paths to avoid negatively affecting one another while pursuing their own goals \cite{silver2005cooperative}. In other cases, agents who collaborate on a joint task must coordinate their plans to efficiently complete the task \cite{ma2016optimal, queralta2020collaborative, dalmasso2021human}.

Optimal multi-agent path planning, even for a centralized planner in a discrete space, is an NP-hard problem \cite{yu2015optimalmultirobotpathplanning}. Path planning for real-world multi-agent cooperation is even more challenging, as agents do not know each other's goals or plans. Humans can solve this problem via communication. We can talk to one another and adjust our plans to cooperate with one another. However, the ability to ground language communication to robot path planning remains challenging for multi-agent cooperation.

 In this work, we formulate robot path planning for multi-agent cooperation as language-guided path editing. As shown in Figure~\ref{fig:concept}, we may edit a robot's path plan based on language communications from other agents (either humans or other robots). This allows the robot to effectively coordinate with a human in both human-robot cooperation, such as joint carrying (Figure~\ref{fig:concept}\textbf{A}) and multi-robot cooperation, such as autonomous driving (Figure~\ref{fig:concept}\textbf{B}). Unlike common paradigms such as replanning \cite{koenig2002d, sharon2015conflict}, editing allows efficient path plan adaptation, which enables flexible adjustment required for many real-world scenarios, such as the example depicted in Figure~\ref{fig:concept}\textbf{A}.
 
To achieve the proposed path editing ability that is both guided by language and grounded in the physical environment, we propose \textbf{\ours}, Code as Path Editor, as illustrated in Figure~\ref{fig:model}. Based on multimodal inputs including the physical state, planner-proposed paths, and the language communication, \ours~leverages a vision-language model (VLM) to generate path editing programs that select and edit from a set of candidate paths proposed by a model-based planner. We further verify the program with the planner to ensure that the program is safe to execute in the physical environment. With the open-endedness of a VLM, \ours~can be flexibly applied to diverse multi-agent scenarios. Unlike prior work that directly generates robot plans with a VLM, \ours~produces path editing programs that can work with a model-based planner to ensure safety. The programs generated by \ours~also enhance the interpretability of robot path planning, allowing humans to more easily understand the success and failure of the robot’s plans. For efficient path editing, we train a small VLM using synthetic data and generalize it to diverse, unseen scenarios, including real-world applications.

As \ours~can take in any physical context, language communication, and planner-proposed path candidates, it can be flexibly integrated into different robotic systems as a plug-and-play module to solve diverse multi-agent cooperation scenarios. These include (1) \textit{multi-robot cooperation} with a varying number of agents communicating with one another and (2) \textit{human-robot cooperation} where the human and the robot either work on separate tasks in a shared environment or jointly perform physical collaboration. In each of these cases, \ours~can align robot path plans with language communication to achieve successful cooperation.

We evaluate \ours~on diverse simulated and real-world multi-agent cooperation scenarios. First, in a simulated parking lot, multiple autonomous vehicles must communicate with one another to negotiate safe paths to avoid collisions. Second, in simulated household environments, a robot assists a human on object arrangement tasks, following human instructions to adapt its path to the human. Third, in the real world, a human and a robot jointly carry an object through tight spaces, where the robot must frequently adjust to the joint plan intended by the human based on their instructions. Across all scenarios, \ours~exhibits strong generalization to unseen layouts, novel tasks, different numbers of agents, and open-ended human instructions, outperforming baselines while maintaining safety and interpretability.

In sum, our main contributions are: (1) a novel framework that uses language-guided program synthesis to edit motion plans while ensuring physical safety through planner-based verification; (2) the integration of \ours~as a plug-and-play module in diverse multi-agent cooperation settings; (3) a VLM finetuning method that enhances a small model's ability to synthesize motion editing programs based on multimodal inputs; (4) new simulated benchmarks for multi-robot and human-robot cooperation focusing on successful path planning guided by language communication; (5) a real-world prototypical system demonstrating the sim-to-real generalization of \ours.

\section{Related Works}
\subsubsection{Program Synthesis for Robots}
Prior works have leveraged VLMs' ability to generate executable code or structured programmatic artifacts to carry out robot tasks via program synthesis. Common approaches include using code as policy \cite{liang2022code, trivedi2021learning}, code as reward functions for task evaluation or learning \cite{venuto2024code, mendez2025systematic, xie2023text2reward, ma2023eureka}, code as symbolic operators for planning \cite{han2024interpret, mahdavi2024leveraging, liu2023llm+}, and code as constraints that parameterize downstream planners \cite{wu2025selp, lin2023text2motion, chen2023nl2tl, pan2023data}. However, existing program synthesis approaches have not been formulated as a general framework for code as a path editor, where synthesized programs explicitly modify an existing plan while preserving interpretability and safety guarantees.

\subsubsection{Communicative Multi-Agent Cooperation}
Prior work on communicative multi-agent systems has primarily focused on using language or learned communication protocols for high-level coordination, such as task allocation, role negotiation, and subgoal planning \cite{zhang2023building,wong2023learning, wan2025infer, tellex2011understanding,qiu2024towards}, while low-level motion generation is handled by separate planning or control modules. In contrast, CaPE focuses on communicative cooperation at the path level, treating language as a direct interface for editing path plans under explicit physical and safety constraints.

\subsubsection{Multi-Agent Path Planning}
Conventional multi-agent path planning assumes a centralized planner that jointly coordinates all agents to avoid collisions, using coupled or constraint-based search in a shared space \cite{van2009centralized, wagner2011m, sharon2015conflict, luna2011efficient}. However, this assumption is restrictive when other agents (e.g., humans or independently-controlled robots) are not part of the centralized optimization loop. Non-centralized alternatives include decentralized collision avoidance that requires no global plan or inter-agent communication \cite{van2011reciprocal, vcap2015prioritized}, as well as decentralized or distributed coordination variants of classical planning approaches. There are also planners that utilize homotopy class constraints \cite{bhattacharya2012topological, kim2012trajectory}. However, these approaches do not ground human language into the planning process, and thus cannot directly leverage natural language instructions or corrections for interactive multi-agent coordination.

\subsubsection{Language-guided Plan Editing}
Several prior works enable language-based plan editing through predicting waypoints, \cite{yuan2024robopoint, lee2025molmoact}, cost functions \cite{sharma2022correcting}, constraints \cite{broad2017real}, direct trajectory modifications \cite{yow2024extract, bucker2022reshaping, bucker2022latte, maurya2025ovita}, modifying the control space \cite{cui2023no}, and using language guidance for robot policy improvement \cite{shi2024yell}. The limitations of these methods include either the lack of interpretability of plan edits or the lack of safety guarantees of the edited plan. In contrast, \ours~produces edits that are easily human-interpretable as the edits consist of simple but comprehensive path-editing operations. The edits are also guaranteed to be collision-free with the environment, as each edit can be cheaply verified by a planner before being applied.

\begin{figure*}[t!]
\centering
\includegraphics[trim=4cm 7cm 4.5cm 7.5cm, clip, width=1.0\textwidth]{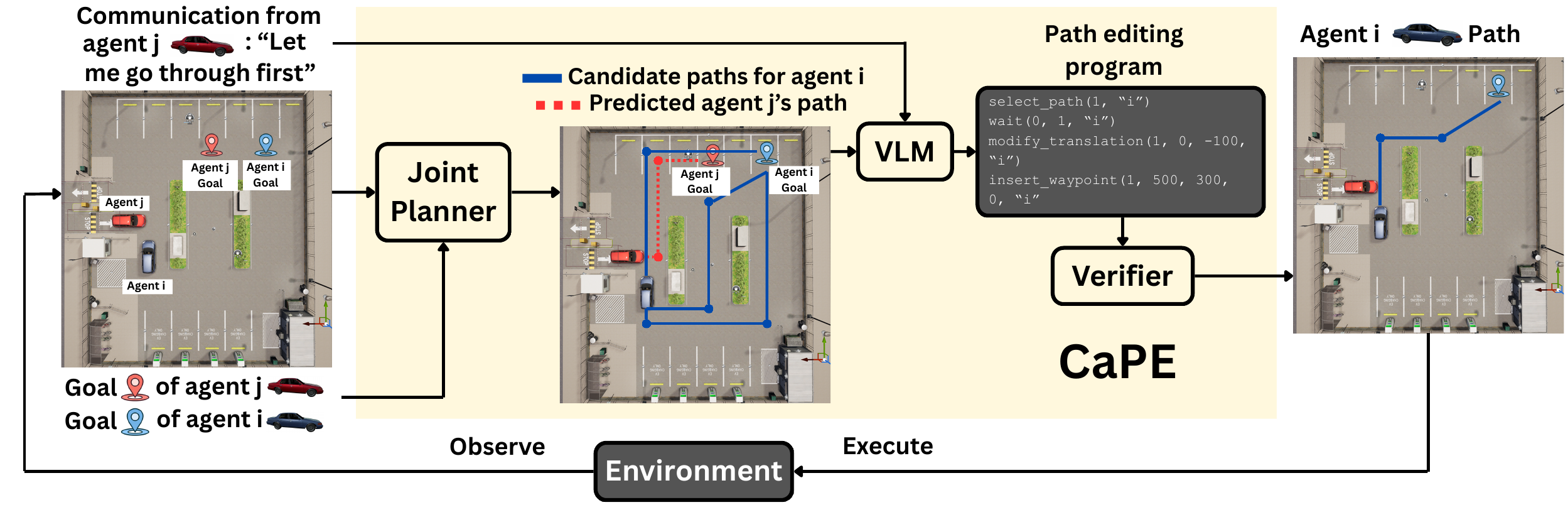}

\caption{Illustration of how \ours~works. The example here shows the target agent $i$'s planning process given language communication from another agent $j$, which can be generalized to scenarios with more agents. From the environment, \ours~takes the current state, the communication, and agent $j$'s goal. It runs the joint planner to plan candidate paths for the target agent and predict a path for the other agent. With the multimodal context, the VLM writes a path edit program, and a verifier checks each line to obtain the final path.}
\vspace{-10pt}
\label{fig:model}
\end{figure*}





\section{Method}

\subsection{\ours: Code as Path Editor}
\subsubsection{Overview} 

As shown in Figure~\ref{fig:model}, \ours~is a general framework for editing an agent's path to align it with the language communication from other agents. Without loss of generality, let us consider the scenarios illustrated in Figure~\ref{fig:model}, where agent $i$ needs to adjust its path to the plan intended by agent $j$'s communication. First, a joint planner simultaneously predicts the path of agent $j$ as well as several candidate paths for agent $i$, based on the observed physical state, agent $i$'s goal, and the goal of agent $j$ (which could be given a priori or predicted by a goal inference model). Conditioned on the paths from the joint planner and the language communication from agent $j$, a VLM generates a path editing program to select and edit one of the candidate paths for agent $i$. Before execution, a verifier based on the planner checks and removes any safety violations in the program and outputs the final, verified path for agent $i$. Note that \ours~can be applied to each agent $i$ in a decentralized way and thus is suitable for both human-robot cooperation and multi-robot cooperation with varying numbers of agents. We discuss each model component in the remaining sections.

\subsubsection{Joint Planner}

To cooperate with other agents, an agent needs to consider a joint plan for all agents. It also needs to imagine multiple candidate paths to account for different ways in which other agents may prefer to act. To this end, we develop a joint planner module for \ours, which (1) proposes multiple candidate paths for the target agent that \ours~is applied to and (2) predicts other agents' paths.

Given the goal of the target agent, the joint planner generates multiple collision-free candidate paths using homotopy-aware path planning \cite{bhattacharya2010search, bhattacharya2012search, liu2023homotopy}. Two paths are said to belong to the same homotopy class if they pass around all obstacles in the same relative order and on the same side. Paths from different homotopy classes correspond to qualitatively different global routes that cannot be transformed into one another through continuous deformation without intersecting obstacles. The planner generates a set of collision-free candidate paths belonging to different homotopy classes by iteratively searching for new solutions and blocking previously discovered classes. We implement this planner using a homotopy-aware variant of RRT \cite{lavalle1998rapidly, liu2023homotopy}. The planner returns a small set of candidate robot trajectories that capture the major global routing options from the start to the goal, each represented as a sequence of waypoints $(x, y, \theta)$ representing the location and orientation of the target agent.

 In addition to the target agent's path, the joint planner also predicts other agents' future paths. Specifically, given the goals of other agents, either given a priori or predicted by a goal inference model, the joint planner computes a collision-free path for each agent via RRT. These predicted paths provide context for the downstream VLM to edit the target agent's path to avoid collision with other agents, ensuring the edited path is compatible with other agents' plans. 

\subsubsection{Path Editing Program Synthesis via a VLM}\label{sec:vlm}
\ours~leverages a VLM to generate a path editing program to select and edit candidate paths for the target agent proposed by the joint planner, conditioned on multimodal context. In particular, the context includes the top-down map constructed from the agent's observation, the candidate paths for the target agent, the predicted paths of other agents, and the language communication from other agents. The objective for the program synthesis is to create a final, adjusted path that is aligned with the communication, satisfies the physical constraints, and avoids conflict with other agents' future paths. A synthesized program first selects the path that best fits the communication. It then further edits the path through operations with waypoints on the paths, including adding new waypoints, modifying the existing waypoints, and adjusting how long the target agent will stay at each waypoint. To achieve this, we define a compact yet comprehensive domain-specific language (DSL) as summarized in Table~\ref{tab:DSL} for the path editing program. 

\begin{table}[t!]
\centering
\begin{tabular}{|p{4cm}|p{4cm}|}
\hline
DSL operation & Functionality \\
\hline
\texttt{select\_path(path\_index, agent)} & select the indexed path \\
\hline
\texttt{modify\_translation(step, dx, dy, agent)} & change the translation of the waypoint at the corresponding step from $(x, y)$ to $(x + dx, y +dy)$ \\
\hline
\texttt{modify\_rotation(step, d$\theta$, agent)} & change the rotation of the waypoint at the corresponding step from $\theta$ to $\theta + d\theta$ \\
\hline
\texttt{wait(step, t, agent)} & stop for time t at the location for the corresponding step \\
\hline
\texttt{insert\_waypoint(step, x, y, $\theta$, agent)} & after the corresponding step insert a waypoint with translation and rotation (x, y, $\theta$) \\
\hline
\end{tabular}
\caption{Our DSL definition.}
\vspace{-5pt}
\label{tab:DSL}
\end{table}

We compile the DSL based on findings from a pilot study detailed in Section V in the supplementary material. In this pilot study, we recruited 10 participants to first perform human-human joint carrying tasks with a PVC pipe and then human-robot tasks where the robot is teleoperated by one of the participants. During the trials, participants gave verbal instructions to their partners or the robot. These instructions include path selection (e.g., ``Let's go from the right''), movement adjustment (e.g., ``move forward a bit into the room, then rotate''), and timing adjustment (e.g., ``let me go first''). These commands are translated to the functions defined in our DSL.

\subsubsection{Training a small VLM as the program synthesizer.} While we can use pretrained VLMs to synthesize the path editing programs, a small VLM finetuned on program synthesis tasks with our DSL can allow \ours~to be run more efficiently for real-world robotic systems without the need for large VLMs, maintaining a low computational cost and latency. For model training, we create a large dataset with diverse program synthesis examples for path editing. In each example, there is a map annotated with candidate paths for the target agent and the predicted paths of other agents, a language instruction, and the ground-truth program. For the training data, we use purely synthetic maps with randomly sampled obstacles. There are diverse layouts and different levels of obstacle cluttering and density. In the training data, we only consider scenarios with two agents. For each map, we sample the starting and goal locations for two agents and synthesize instructions and the corresponding programs using templates. We further diversify the style of the instructions via a large language model (LLM). In total, there are 50k training examples in the dataset. Section IV provides more details of the data generation and model finetuning. 

\subsubsection{Path Verifier}

To ensure safety, \ours~verifies all edits generated by the path editing program before execution. Unlike usual programs, each operation defined in our DSL defines a standalone modification to the path that functions without depending on previous lines. The only exception is path selection. However, since it is always executed at the beginning of the program and the path it selects is from a planner, its outcome will never result in constraint violation. Given this, the verifier in \ours~goes through each line of code's intended path edit with the planner to check against the constraints imposed by the physical environment and other agents' future paths. It then rejects the lines introducing edits that will violate the constraints. The revised program will thus be safe for the target agent to execute. There may be cases where the revised program will produce an edited path that slightly deviates from the path intended by the communication. However, this deviation can be mitigated in two ways. First, a strong VLM (either a large VLM with a strong multimodal reasoning capacity or a finetuned VLM as discussed in Section~\ref{sec:vlm}) significantly reduces both the frequency of producing code that will cause violations and the needs of major program revision that may lead to a large deviation. Second, future communications can also adjust the deviation.



\subsection{Integration of \ours~in Multi-Agent Cooperation}\label{sec:integration}

\begin{figure*}[t!]
\centering
\includegraphics[trim=0cm 14cm 13.2cm 0cm, clip, width=1.0\textwidth]{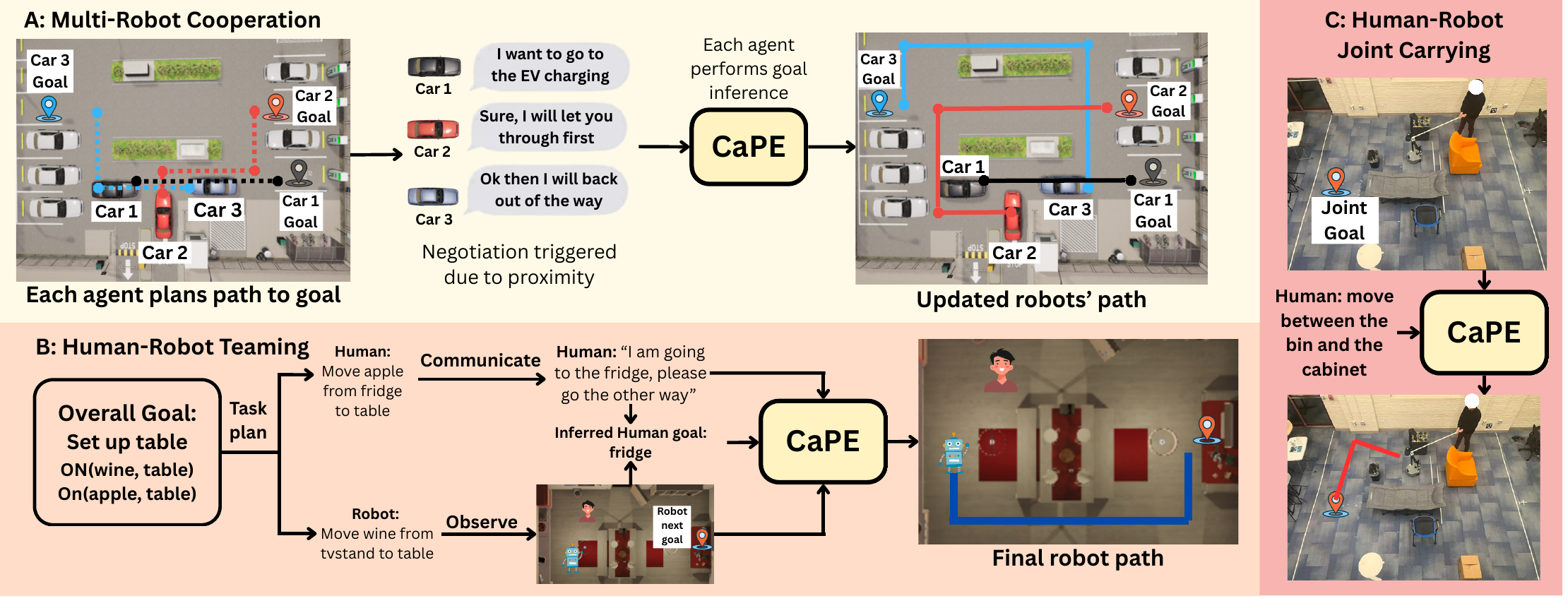}

\caption{Representative examples of multi-agent cooperation scenarios and the corresponding \ours~integration. (\textbf{A}) In multi-robot path coordination, it can take in the communication from the negotiation among multiple robots and edit the path for all robots accordingly. (\textbf{B}) In human-robot teaming for household object rearrangement, it can adjust the robot's low-level path given high-level task assignments for the human and the robot, as well as the human instructions. (\textbf{C}) In human-robot teaming for joint carrying, it can align the robot's path with the one intended by the human instructions.}
\vspace{-5pt}
\label{fig:integration}
\end{figure*}

\ours~is designed as a plug-and-play module that can be applied to different robotic systems across diverse multi-agent cooperation scenarios. Figure~\ref{fig:integration} provides three representative examples of how \ours~can be integrated into multi-agent cooperation scenarios, which we discuss in detail below.

\subsubsection{Multi-Robot Path Coordination}
Figure~\ref{fig:integration}\textbf{A} illustrates how CaPE can be integrated into multi-robot collaboration tasks. Each agent can negotiate a safe path with other agents when they get close. Based on the negotiation, each agent predicts the goals of the other agents, and then runs CaPE to edit its path to reflect the outcome of the negotiation and safely reach its intended goal.

\subsubsection{Human-Robot Teaming for Household Object Rearrangement}

Figure~\ref{fig:integration}\textbf{B} illustrates how \ours~can be integrated into human-robot teaming tasks for household object rearrangements. In such tasks, the human and the robot complete complementary tasks for an overall goal. With high-level task planning, each agent is assigned a set of tasks (i.e., a set of objects to rearrange in this case). From the environment, the robot can observe the current state and track the task progress to get the next location to go to. The human may also instruct the robot so that it can better adapt its behavior to the human's preference. Given the human instruction and the current state, the robot can infer the next goal location of the human. \ours~can integrate the multimodal context and plan a robot path aligned with human instructions while completing its assigned tasks.

\subsubsection{Human-Robot Teaming for Joint Carrying}

Figure \ref{fig:integration}\textbf{C} illustrates how \ours~can be integrated into human-robot joint carrying tasks. In these tasks, a robot can observe the environment to get the current state and receive language instructions from the human about specific ways to get to the goal location. In this case, the goal location is given to both the human and the robot a priori. With the multimodal context, \ours~can be used to plan a path for the robot to physically collaborate with the human to carry the object to the goal location in a way preferred by the human.

\section{Experiments}

To test \ours's generalizability to different multi-agent cooperation scenarios, we designed different evaluation domains inspired by the three representative examples as discussed in Section~\ref{sec:integration}. They present different challenges for robot path planning in multi-agent cooperation: multi-robot coordination requires multiple robots to avoid colliding with each other without knowing each other's intentions or plans; human-robot teaming tasks in household environments require robots to navigate and avoid collision in a tight space with dense obstacles; and human-robot teaming tasks with joint-carrying require motion level alignment to the human's intended path. They are also out of distribution relative to our training data: multi-robot coordination involves outdoor objects and multiple agents that never appear in the training data; household environments have different layouts and a higher density of objects than the training maps; and real-world joint-carrying tasks test the sim-to-real gap using real human instructions that differ significantly from the templated instructions used during training.

\subsection{Experiments 1: Multi-Robot Path Coordination for Autonomous Vehicles}

\subsubsection{Setup}
We evaluated our models on a set of carpark coordination scenarios for autonomous vehicles in a photorealistic embodied outdoor simulator, SimWorld~\cite{yesimworld, zhuang2025simworld}, which reflects common interactions in real-world parking lots. In these scenarios, either 2 or 3 cars need to navigate from their starting positions to their goal positions without colliding with any obstacles or other cars. This must be accomplished without knowing the other cars' goals or plans ahead of time. The cars must negotiate with one another and edit their paths accordingly to avoid collisions. The 2-car settings reflect challenging cases where the cars start close together, and the 3-car settings have randomly sampled start and goal locations.
\subsubsection{Baselines}

We implemented the following baselines: (1) Planner-only, where the agents run a RRT-based planner to execute the task without any path update based on the communication; (2) VLM agents, where VLMs are prompted with top-down \& egocentric views of the map to generate actions within the action space \texttt{[move\_forward, move\_backward, turn\_left, turn\_right, stay]} to reach the goal; (3) VLM waypoint Generators, where VLMs are prompted with top-down views to directly generate a path; and (4) Robopoint \cite{yuan2024robopointvisionlanguagemodelspatial}, a VLM trained on robotic tasks like affordance finding or navigation. We tested the following VLMs: Gemini-3-pro, \cite{deepmind_gemini3pro_2025}, GPT 5.2 \cite{openai_gpt5_2_2025}, Claude Opus 4.5 \cite{anthropic_claude_opus45_2025}, Qwen 2.5 72B (larger version of the base model for our training), and Molmo 8B \cite{deitke2024molmopixmoopenweights} (a VLM trained for spatial reasoning) as VLM agents, VLM waypoint generator, or path editing program synthesizer for \ours. We also fine-tuned a Qwen 2.5 7B model as the motion program synthesizer with synthetic maps (refer to section IV in the supplementary material for more information), and also tested our finetuned version and the base model as the path editing program synthesizer for \ours. We ran all baselines for 2-robot experiments, and only evaluated the best-performing model under each category for 3-robot experiments.

We reported the following metrics for the experiments: success rate (SR), with failure occurring when agents collide into each other or with obstacles; inference (inf) and token cost for the model inference; and Success Efficiency Length (SEL), which quantifies the efficiency of the agent's actions by comparing the ratio of the optimal time to the actual time taken to complete each task.

\subsubsection{Results}

\begin{table}[t!]
\centering
\caption{\textbf{Multi-robot coordination results with 2 robots (IV.A).}
\ours~achieves significantly higher success rates than planner-only and VLM-based baselines, enabling effective multi-agent coordination where conventional planning and waypoint generation fail.}
\setlength{\tabcolsep}{3pt}
\begin{tabular}{lp{1cm}p{1cm}p{1cm}p{1cm}}
\toprule
\textbf{Models} &
\textbf{SR $\uparrow$} &
\textbf{SEL $\uparrow$} &
\textbf{Time(s)$\downarrow$} &
\textbf{Token$\downarrow$} \\
\midrule
Planner-only & 23.3 & 15.1 & 0 & \textbf{0} \\
RoboPoint & 0 & 0 & 32.8 & 2266 \\
VLM Agents &&&&\\
\quad w/ Gemini 3.0 pro & 0 & 0 & 44.1 & 3477. \\
\quad w/ Claude Opus 4.5 & 0 & 0 & 12.7 & 2807 \\
\quad w/ GPT 5.2 & 10.0 & 0.6 & 56.5 & 2828 \\
\quad w/ Molmo 8B & 0 & 0 & 32.8 & 2266 \\
\quad w/ Qwen 72B & 0 & 0 & 22.0 & 2540 \\
Waypoint &&&&\\
\quad w/ Gemini 3.0 pro & 16.6 & 0.1 & 95.5 & 3902 \\
\quad w/ Claude Opus 4.5 & 10.0 & 0.2 & 28.7 & 2771 \\
\quad w/ GPT 5.2 & 20.0 & 0.5 & 39.6 & 2677 \\
\quad w/ Molmo 8B & 0 & 0 & 10.3 & 2042 \\
\quad w/ Qwen 2.5 72B & 0 & 0 & 34.5 & 2383 \\
\ours~&&&&\\
\quad w/ Gemini 3 pro & 76.6 & 5.9 & 50.5 & 1642 \\
\quad w/ Claude Opus 4.5 & 73.3 & 6.7 & 42.5 & 1610 \\
\quad w/ GPT 5.2 & 76.6 & 5.0 & 43.0 & 1625 \\
\quad w/ Molmo 8B & 46.6 & 3.6 & 42.1 & 1111 \\
\quad w/ Qwen 2.5 7B & 73.3 & 5.8 & 41.6 & 1832 \\
\quad w/ Qwen 2.5 72B & 73.3 & 6.0 & 39.1 & 1593 \\
\quad w/ Finetuned Qwen 2.5 7B & \textbf{90.0} & \textbf{18.0} & \textbf{7.0} & 1623 \\
\bottomrule
\end{tabular}
\label{tab:multi-2}
\end{table}

\begin{table}[t!]
\centering
\caption{\textbf{Multi-robot coordination with 3 robots (IV.A).}
\ours~maintains strong performance as the number of robots increases, achieving up to 60.0\% success rate, while VLM agent and waypoint baselines fail on all tasks.}
\setlength{\tabcolsep}{3pt}
\begin{tabular}{lp{1cm}p{1cm}p{1cm}p{1cm}}
\toprule
\textbf{Models} &
\textbf{SR $\uparrow$} &
\textbf{SEL $\uparrow$} &
\textbf{Time(s)$\downarrow$} &
\textbf{Token$\downarrow$} \\
\midrule
Planner-only & 30.0 & 25.9 & 0 & 0 \\
VLM Agents (w/ Gemini 3 pro) & 0 & 0 & 33.9 & 4125 \\
Waypoint (w/ Gemini 3 pro) & 0 & 0 & 14.8 & 3666 \\
\ours~(w/ Gemini 3 pro) & \textbf{60.0} & 8.9 & 43.5 & 1426 \\
\ours~(w/ Finetuned Qwen 2.5 7B) & 50.0 & \textbf{18.8} & \textbf{5.9} & 1257 \\
\bottomrule
\end{tabular}
\label{tab:multi-3}
\end{table}

Tables \ref{tab:multi-2} and \ref{tab:multi-3} show the results for multi-robot collaboration among 2 robots and 3 robots, respectively. For the 2-robot cases, all the VLM baselines fail spectacularly, with the best success rate of Gemini-3-pro generating waypoints only $16.67$\%. \ours~with different VLMs all showed significant improvements against all baselines. Planner-only baselines can succeed in only 23.33\% of the trials, suggesting that our evaluation tasks are challenging and cannot be solved by just planning. VLMs trained for robotic tasks for spatial reasoning, like Molmo and Robopoint, still fail to solve the coordination task effectively. With our training, the finetuned model achieved the best performance, suggesting generalizability to outdoor maps. Testing the best-performing VLM Gemini-3-pro in 3-robot cases, we find that \ours~is able to effectively generalize to more agents, achieving a 60\% success rate when other VLM-based baselines fail all the tasks. Our finetuned model shows a significant drop in performance while still being comparable to larger VLMs, suggesting challenges for generalizing to the 3-robot case.

\subsection{Experiment 2: Human-Robot Teaming for Household Object Rearrangement}

\subsubsection{Setup}
We evaluated our models on a set of human-robot teaming scenarios for household object rearrangement in a household simulator, VirtualHome 2.0 \cite{puig2018virtualhome, puig2020watchandhelp}, which reflects common coordination patterns in everyday household environments. Our setup simulated situations where the human and the robot have completed task-level planning while still requiring path-level coordination. We generated 20 rearrangement scenarios with fixed sub-task assignments for the human and the robot across 7 different apartments, each involving human and robot acting together across multiple rooms. In these scenarios, a simulated human follows a pre-planned path and generates instructions when approaching the robot. The robot must align with the human and avoid collisions. We run all the baselines from Experiment 1 for comparison.

\subsubsection{Results}

Table \ref{tab-team} summarizes the results of the human-robot teaming task in VirtualHome household environments. Still, the coordination tasks are challenging to solve just by the planner. In the multi-room environment, all VLM agents failed for all the tasks. With more cluttered objects and tighter spaces in apartments than in the parking lot, VLM Waypoints perform much worse, with the highest success rate being only 8.33\%. Robopoint, specifically trained for robotic tasks, fails all tasks in the multi-room household environment. \ours, combined with different VLMs, all show significant improvement against the baselines, with the best performing model, Gemini-3-pro, succeeding in 80\% of the tasks. The fine-tuned Qwen 2.5 7B model offers fewer improvements for the household tasks, probably due to an increase in obstacle density and a more significant change in layouts compared to the training data, while still showing improvement against the base model and being comparable to the larger 72B model.

\begin{table}[t!]
\centering
\caption{\textbf{Human-robot teaming in VirtualHome household environments (IV.B).}
\ours~significantly improves success rate and SEL compared to planner-only, RoboPoint, and VLM-based baselines. The best-performing model (Gemini 3.0 pro) achieves 80.0\% success rate, while VLM agent baselines fail and waypoint baselines achieve at most 8.3\% success rate.}
\setlength{\tabcolsep}{3pt}
\begin{tabular}{lp{1cm}p{1cm}p{1cm}p{1cm}}
\toprule
\textbf{Models} &
\textbf{SR $\uparrow$} &
\textbf{SEL $\uparrow$} &
\textbf{Time(s)$\downarrow$} &
\textbf{Token$\downarrow$} \\

\midrule
Planner-only & 8.3 & 4.9 & 3.2 & \textbf{0} \\
RoboPoint & 3.3 & 0.8 & 2.1 & 507 \\ 
VLM Agents & & & & \\
\quad w/ Gemini 3.0 pro & 0.0 & 0.0 & 17.0 & 4238 \\
\quad w/ Claude Opus 4.5 & 0.0 & 0.0 & 13.7 &  2378 \\
\quad w/ GPT5.2 & 0.0 & 0.0 & 2.2 & 2195 \\
\quad w/ Molmo 8B & 0.0 & 0.0 & 2.2 & 2175\\
\quad w/ Qwen 2.5 72B & 0.0 & 0.0 & 3.0 & 2307 \\
VLM Waypoints & & & & \\
\quad w/ Gemini 3 pro & 8.3 & 2.1 & 4.6 & 1794 \\
\quad w/ Claude Opus 4.5 & 1.6 & 0.4 & 12.4 & 1790 \\
\quad w/ GPT5.2 & 3.3 & 1.4 & 2.3 & 1673 \\
\quad w/ Molmo 8B & 0.0 & 0.0 & 6.0 & 1827 \\
\quad w/ Qwen 2.5 72B & 5.0 & 0.7 & 3.2 & 1539 \\
\ours~& & & & \\
\quad w/ Gemini 3.0 pro & \textbf{80.0} & \textbf{22.5} & 8.8 & 2711 \\
\quad w/ Claude Opus 4.5 & 70.0 & 20.9 & 6.8 & 2306 \\
\quad w/ GPT 5.2 & 65.0 & 18.2 & 10.1 & 2539 \\
\quad w/ Molmo 8B & 55.0 & 19.0 & 1.7 & 1740 \\
\quad w/ Qwen 2.5 7B & 41.6 & 17.1 & \textbf{1.1} & 982 \\
\quad w/ Qwen 2.5 72B & 46.6 & 18.0 & 1.9 & 2304 \\
\quad w/ Finetuned Qwen 2.5 7B & 48.3 & 16.5 & 1.7 & 2448 \\

\bottomrule
\end{tabular}
\label{tab-team}
\end{table}

\subsection{Experiment 3: Real-World Human-Robot Joint Carrying}

\subsubsection{Setup}
\begin{figure}[t!]
\centering
\includegraphics[trim=0cm 24cm 17cm 0cm, clip, width=0.9\textwidth]{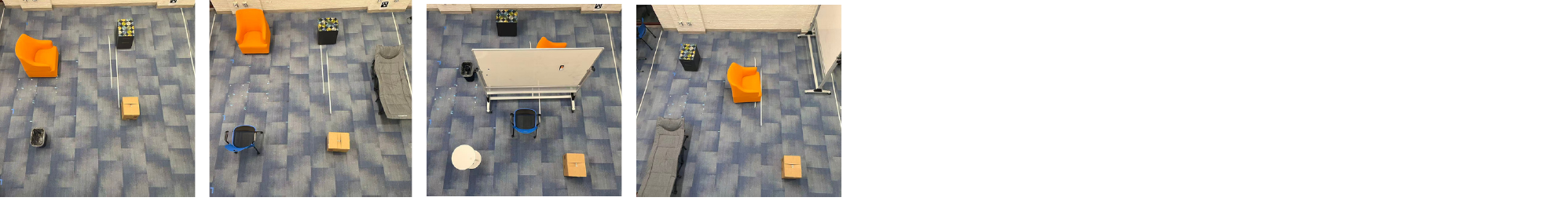}
\caption{Four maps we designed for human and robot joint carrying experiments. The maps are used with common office objects in an open space. For each map, the human and the robot start at the right of the map and need to get to the goals in the left region.}
\vspace{-5pt}
\label{fig:map}
\end{figure}

To evaluate the real-world generalization of \ours, we tested it on real-world human-robot joint carrying tasks. In this experiment, a human and a Stretch 3 robot \cite{kemp2022designstretchcompactlightweight} carry a PVC pipe together in a tight space with multiple obstacles. Successful task completion in this experiment requires smooth maneuvering around the obstacles without any collision. The total space is 6 meters $\times$ 6 meters, with 4 to 7 obstacles arranged in different layouts (as shown in Figure~\ref{fig:map}). This setup is significantly more challenging than the conditions tested in recent work on human-robot joint carrying \cite{yang2025implicitcommunicationhumanrobotcollaborative}.


We recruited 10 participants (average age = 22.6, 4 females, 6 males) from a college to perform our tasks. Each participant consecutively performed 4 joint-carrying tasks on 4 maps with a Stretch 3 robot driven by 4 different methods: \ours~with Gemini-3-pro and our finetuned Qwen 2.5 7B model, Gemini-3-pro waypoint prediction (the best performing baseline in simulation), and the planner-only baseline. To reduce any bias for participants caused by the order of the sessions or the difficulty of the maps, we randomized the order of the models and the maps for each participant. We reported the following metrics for the experiments: success rate, runtime, and token cost.

After each task, we ask the participant to rate the collaboration in the following dimensions: instruction comprehension, comfortable human movements, efficiency, responsiveness, safety (exact prompts are provided in the section III of the supplementary material) with a Likert scale. Following previous works that collect human preferences for more than 2 methods \citep{jain2013learningtrajectorypreferencesmanipulators,palan2019learningrewardfunctionsintegrating}, we ask the participant to rank the 4 models without any ties. All the experiments are approved by an Institutional Review Board (IRB). 

\subsubsection{Result}

\begin{figure}[t!]
\centering
\includegraphics[trim=0cm 0cm 0cm 0cm, clip, width=0.48\textwidth]{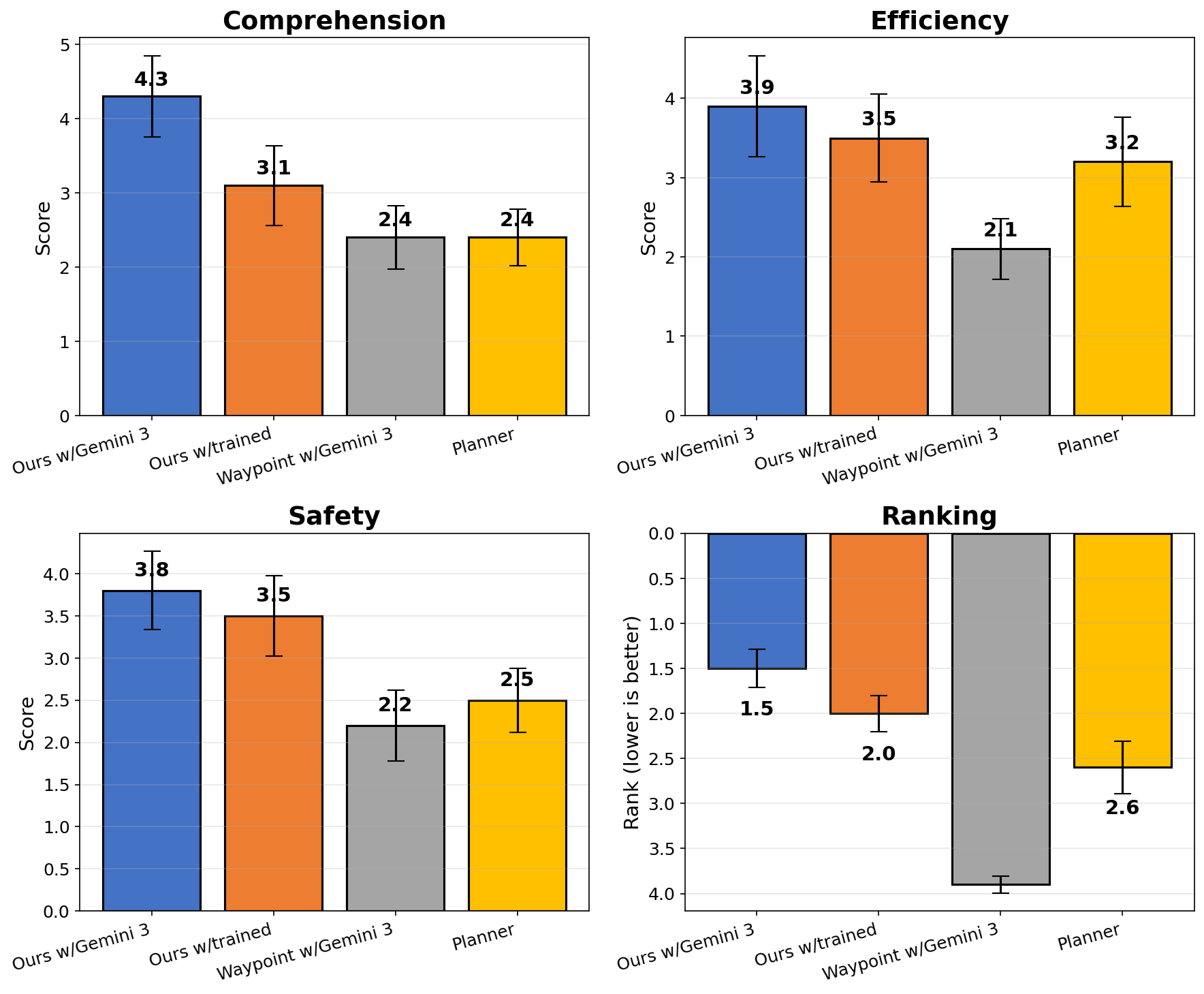}
\caption{Participants' subjective ratings for the models. Participants rated how well the robot comprehended their instructions, how efficient the robot's selected path was, and how safe the overall execution was for both the robot and the human. They also ranked their collaboration with the four models. The numbers indicate means; the error bars indicate standard errors.}
\label{fig:rating}
\end{figure}

Table \ref{tab:real} shows the results of the real-robot experiment. Compared with planner-only and the best baseline Gemini-3-Pro Waypoint in simulation, both \ours~with Gemini 3 (also the best performing) and our trained model significantly outperform them, succeeding 7 of the 10 trials. Figure \ref{fig:rating} shows the barplots of participants' subjective evaluation of the collaboration. \ours~with Gemini-3 outperforms the Planner baseline in comprehension, safety, and ranking. Our trained model outperforms the planner baseline by a smaller margin, but infers 2 times faster, suggesting that our trained model can generalize to real-world instructions and is more efficient. For Gemini Waypoint generation, it fails almost all the trials in the real-world. 
\begin{table}[t!]
\centering
\caption{\textbf{Real-world human-robot joint carrying (IV.C).}
Both variants of \ours~achieve a higher success rate (70\%) than planner-only (20\%) and waypoint baseline (10\%). The finetuned model achieves lower runtime than the Gemini-based variant while maintaining the same success rate.}
\setlength{\tabcolsep}{3pt}
\begin{tabular}{lp{1cm}p{1cm}p{1cm}}
\toprule
\textbf{Models} &
\textbf{SR $\uparrow$} &
\textbf{Time(s)$\downarrow$} &
\textbf{Token$\downarrow$} \\
\midrule
Planner-only & 20 & 9.4 & \textbf{0} \\
Waypoint (w/Gemini 3 pro) & 10 & \textbf{5.8} & 1962 \\
\ours~(w/ Gemini 3 pro) & \textbf{70} & 32.7 & 7694 \\
\ours~(w/ Finetuned Qwen 2.5 7B) & \textbf{70} & 14.3 & 7281 \\
\bottomrule
\end{tabular}
\label{tab:real}
\end{table}

\begin{figure}[t!]
\centering
\includegraphics[trim=0cm 14cm 10cm 0cm, clip, width=1.0\textwidth]{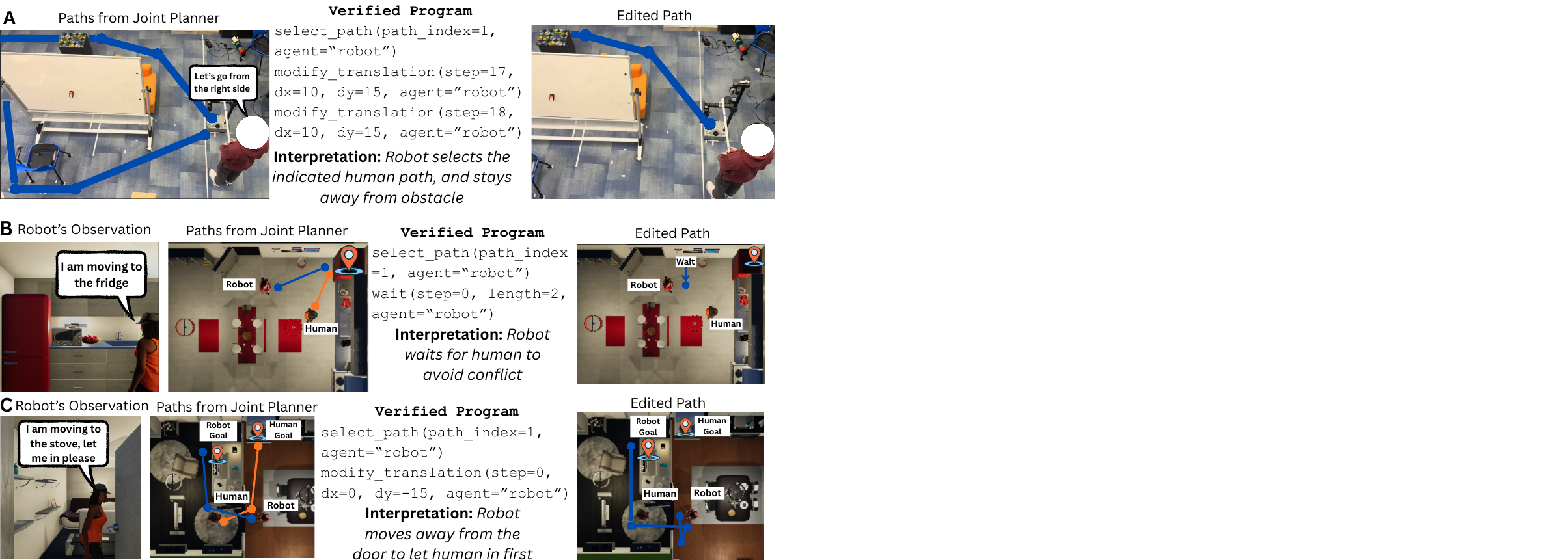}
\caption{Qualitative examples of \ours~path modification effect. (A) In a real-world joint-carrying task, the path editing program selects the human's intended path and avoids colliding with obstacles. (B) In a household rearrangement task, when the robot and the human are both approaching the fridge, the robot waits for the human to get to the fridge first. (C) In a household rearrangement task, when the robot and the human are passing through a narrow door, the robot moves away to let the human in first. The path from the joint planner is showing candidate paths for the robot and the predicted path for the human.}
\vspace{-10pt}
\label{fig:qual}
\end{figure}

\section{Discussion}

\subsection{\ours~provides safe and interpretable language-based plan edit}

Across all experiments, \ours~achieves significantly higher success rates than all baselines. Most baseline failures arise from generating physically invalid paths, such as placing waypoints inside obstacles or too close to obstacle boundaries. In contrast, \ours~edits planner-generated paths through validated, constraint-aware modifications, ensuring that all returned paths satisfy physical feasibility requirements.

In simulation, \ours~never produces paths that collide with obstacles or map boundaries. In real-world deployment, execution and perception noise introduce additional uncertainty. Although both \ours~and the planner generate physically valid paths in computation, these paths may still fail during execution due to localization and control noise. Despite these challenges, \ours~maintains a substantially higher success rate. Its edits increase clearance from obstacles, making the resulting paths more robust to real-world noise and reducing execution failures.

We further analyze the generated edit programs. Across all trials, both Gemini-3-Pro and our trained model frequently modify waypoint positions. Gemini-3-Pro often adjusts waypoint locations even when human instructions specify only high-level directional guidance, proactively increasing distance from nearby obstacles (see Figure \ref{fig:qual}). This behavior occurs in 8 out of 10 participants when using Gemini within \ours. In contrast, our trained model more reliably interprets fine-grained instructions, such as ``move backward a little," enabling precise local adjustments that improve obstacle clearance. In 6 out of 10 trials, the trained model correctly applies these local movement instructions to maintain safe distances.

Figure \ref{fig:qual} presents several representative examples of successful edits produced by \ours. In Figure \ref{fig:qual}A, the generated program selects the right-most path from the proposed candidates and proactively applies multiple edits to increase clearance from nearby obstacles, despite the absence of explicit human instructions. This demonstrates the model’s ability to improve safety through refinement of planner outputs.

Figures \ref{fig:qual}B and \ref{fig:qual}C illustrate two examples in indoor household environments, where the robot operates in tight shared spaces. In these scenarios, the generated programs enable effective coordination by modifying the robot’s motion to accommodate human movement. For example, the robot chooses to wait or adjusts its position away from the doorway, ensuring safe and efficient passage.

Beyond improving navigation safety and coordination, \ours~is inherently interpretable. Each edit is expressed as a well-defined DSL operation applied to specific waypoints, allowing the edits to be directly grounded in natural language intent. For instance, in Figure \ref{fig:qual}B, the wait operation at step 0 corresponds to the robot pausing at its current location to allow the human to pass first. Similarly, in Figure \ref{fig:qual}C, modifying the first waypoint causes the robot to move out of the doorway, clearing space for the human.

Because edits are expressed as explicit motion-editing programs, users can directly inspect how the path was modified and interpret the robot’s coordination strategy.


\subsection{\ours~generalizes well to diverse scenarios}

Our evaluation spans diverse scenarios, including indoor, outdoor, and real-world tasks, each featuring different layouts and obstacle types. For outdoor simulations, we use a parking lot with objects such as fences. Indoor simulations consist of multi-room apartment layouts with dense objects. All of these environments are out of distribution relative to our training data. We use different VLMs as motion program synthesizers, and \ours~consistently achieves strong performance in cooperation tasks, suggesting generalizability across layouts and coordination scenarios. The three-robot coordination experiments further show that \ours~is able to generalize to a larger number of agents.

Although our trained model is trained only on two-agent cases and synthetic maps, it generalizes well to unseen objects and layouts and performs best in the two-robot coordination task. However, its generalization is weaker in the three-robot and indoor tasks.

\ours~also demonstrates advantages in real-world settings. Participants provide highly diverse instructions, using different units to describe motion, different reference perspectives, and varied phrasing for similar intents. With both Gemini-3-Pro and the trained model, \ours~shows the ability to comprehend diverse instructions from different users.

\subsection{Limitations and future work}
The input to \ours~assumes reasonably reliable visual perception of the scene. In real-world settings, however, perception noise and partial observability may introduce challenges, which can affect the quality of the generated path edits. Incorporating more robust perception modules or uncertainty-aware representations could help improve performance under such conditions.

In addition, \ours~relies on predicted goals and trajectories of other agents. Errors in these predictions, especially in human-robot scenarios where intentions may evolve over time, can lead to suboptimal edits. While such errors are difficult to fully avoid, the interactive nature of \ours~provides a mechanism for mitigation through additional communication and subsequent plan updates, allowing the agent to gradually refine its behavior as new information becomes available.

Additionally, although \ours~enforces safety through planner-based verification, these safety checks reject infeasible edits, and the resulting plan may not always exactly reflect the human's original intent. This reflects a trade-off between strict safety enforcement and faithful instruction execution.

Finally, \ours~is not intended to learn collective policies for large groups of agents. Instead, it focuses on enabling an individual agent to flexibly adapt its behavior when interacting with decentralized partners or humans through communication. Future work could extend the framework to richer forms of multimodal communication, such as gestures or implicit signals, and improve robustness under perception uncertainty.

\section{Conclusion} 
We presented CaPE, a safe and interpretable framework for multimodal path editing in multi-agent cooperation. CaPE leverages vision-language models to synthesize path-editing programs and verifies each edit with physical constraints, enabling agents to adapt their behavior to decentralized partners or humans while maintaining geometric safety constraints.

Unlike prior approaches that map language directly to actions or trajectories, CaPE treats communication as guidance for editing existing plans, allowing for efficient, incremental adaptation and transparent reasoning about editing decisions. Through extensive experiments in simulated multi-robot coordination, human-robot teaming, and real-world joint carrying, we demonstrate that CaPE significantly improves cooperation performance over strong baselines while preserving safety and interpretability. We believe CaPE represents a step toward a broader paradigm of interactive robot coordination, in which natural language is used to guide verifiable path edits rather than directly generating paths or low-level actions.


\bibliographystyle{plainnat}
\bibliography{references}

\newpage

\section{Appendix}

\subsection{More Results}

To better understand the effect of using a homotopy-aware planner as well as the safety check for \ours, we perform two ablation studies.

\subsubsection{Ablation 1: \ours~using single path planner}
In this ablation, we evaluate \ours~using a planner that generates only a single collision-free robot path, without homotopy-based planning. We use GPT 5.2 as the VLM for the Multi-Robot Coordination task and Gemini 3 Pro as the VLM for the Human-Robot Teaming task, as these models achieved the highest performance on their respective tasks. Table \ref{tab:ablation-1} shows the results. The single-path planner achieves lower performance in both multi-robot coordination and Human-Robot teaming scenarios. Notably, the performance drop is steeper for Human-Robot teaming, likely because the carpark map is more spacious than the indoor environments in VirtualHome.

\begin{table}[h!]
\centering
\caption{Results of \ours~using single path planner.}
\setlength{\tabcolsep}{3pt}
\begin{tabular}{lp{1cm}p{1cm}p{1cm}p{1cm}}
\toprule
\textbf{Models} &
\textbf{SR $\uparrow$} &
\textbf{SEL $\uparrow$} &
\textbf{Time(s)$\downarrow$} &
\textbf{Token$\downarrow$} \\
\midrule
Multi-Robot Coordination & & &  & \\
\quad w/ Homotopy-Aware Planner & 55.0 & 4.9 & 41.0 & 1552\\
\quad w/ Single-Path Planner & 53.4 & 6.7 & 37.2 & 1504\\
Human-Robot Teaming & & & & \\
\quad w/ Homotopy-Aware Planner & 80.0 & 22.5 & 8.8 & 2711\\
\quad w/ Single-Path Planner & 63.3 & 17.0 & 5.0 & 2249\\
\bottomrule
\end{tabular}
\label{tab:ablation-1}
\end{table}

\subsubsection{Ablation 2: \ours~without safety check}
In this ablation, we evaluate \ours~without feasibility checks for the edits, which removes safety guarantees for the output paths. We use GPT 5.2 as the VLM for the Multi-Robot Coordination task and Gemini 3 Pro as the VLM for the Human-Robot Teaming task, as these models achieved the highest performance on their respective tasks. Table \ref{tab:ablation-2} shows the results. Similar to Ablation 1, removing safety checks in \ours~leads to slightly lower performance. The larger performance drop in VirtualHome is likely due to the more cluttered maps, as VLMs may be more prone to generating modifications that result in collisions.

\begin{table}[h!]
\centering
\caption{Results of \ours~without feasibility checks for the edits}
\setlength{\tabcolsep}{3pt}
\begin{tabular}{lp{1cm}p{1cm}p{1cm}p{1cm}}
\toprule
\textbf{Models} &
\textbf{SR $\uparrow$} &
\textbf{SEL $\uparrow$} &
\textbf{Time(s)$\downarrow$} &
\textbf{Token$\downarrow$} \\
\midrule
Multi-Robot Coordination & & &  & \\
\quad w/ Safety Check & 55.0 & 4.9 & 41.0 & 1552\\
\quad w/o Safety Check & 53.4 & 7.0 & 42.7 & 1538\\
Human-Robot Teaming & & & & \\
\quad w/ Safety Check & 80.0 & 22.5 & 8.8 & 2711\\
\quad w/o Safety Check & 60.0 & 14.7 & 7.3 & 2440\\
\bottomrule
\end{tabular}
\label{tab:ablation-2}
\end{table}

\subsection{\ours~Implementation Details}

\subsubsection{Prompts for VirtualHome Evaluation}

\begin{tcolorbox}[breakable]
You are a motion planning assistant who selects and modifies waypoint-based paths. The map is a representation of a human and a robot acting separately in a shared space. The keypoints indicated the location and rotation of the robot and humans. You will receive a list of the human and the robot's waypoints in text format, an image of the map, and some instructions. You should select and modify the path in a way that complies with the current scenario and the instructions. \\

AVAILABLE TOOLS: \\
1. select\_path(path\_index, agent): Select a path for the agent by index. Path should be 0-indexed. \\
2. modify\_translation(step, dx, dy, agent): Change the specified agent (human or robot)'s position by dx, dy at the step. \\
3. modify\_rotation(step, degrees, agent): Change the agent's rotation at the step by degree (+ indicates clockwise, - indicates counterclockwise) (normalized to [-180, 180]). The waypoint's rotation is with respect to the world. Rotation 0° indicates the human or the robot facing the right. \\
4. wait(step, length, agent): After the waypoint at the indicated step, have the agent stay at the same point for length steps, then execute the left waypoints. \\
5. insert\_waypoint(step, x, y, agent): insert a new waypoint after the step at position x, y for the agent. \\

The image shows: \\
- Lines of different colors with legends: Human and robot paths. \\
- Brown blocks with names: Obstacles with their name. \\
- Grey blocks: Unreachable regions. \\

IMPORTANT:
- First call select\_path() for the planning agent. \\
- All the other code should assume that each agent only has one path for now. \\
- Only use the operations above (no other variables/operations). \\
- Use obstacle names and their relative positions for spatial reasoning. \\
- Right is +x, down is +y for coordinates. \\
- You should try to wait or adjust your plan if and only if your plan seems to interfere with the human trajectory. \\
- Do not generate unnecessarily long waiting codes, you should still try to finish your own task efficiently in a collision free way. \\
- Always use the name "robot\_1" as the name for the agent. Don't use other things. \\

Return only the function code (no extra text).

\end{tcolorbox} 

\subsubsection{Prompts for SimWorld Evaluation}

\begin{tcolorbox}[breakable]
\textbf{Prompt for DSL Generation}\\
Convert this instruction into DSL code for \{target\_agent\}.\\

INSTRUCTION: \{instruction\}\\

\{target\_agent.upper()\}'S CANDIDATE PATHS (0-indexed):\\
\{paths\_text\}\\

OTHER AGENTS (reference only):
\{other\_agents\_text\}\\

DSL OPERATIONS (for \{target\_agent\} only):\\
1. select\_path(path\_index, ``\{target\_agent\}") - Choose path by index (0, 1, or 2). Path 0 is the first path shown.\\
2. wait(step, length, ``\{target\_agent\}") - Pause at waypoint\\
3. modify\_translation(step, dx, dy, ``\{target\_agent\}") - Move ONE waypoint\\
4. modify\_rotation(step, degrees, ``\{target\_agent\}") - Rotate ONE waypoint\\
5. insert\_waypoint(step, x, y, ``\{target\_agent\}") - Add waypoint\\

RULES:
- Output ONLY Python code\\
- One operation per line\\
- Only modify \{target\_agent\}\\
- IMPORTANT: Path indices are 0-based (Path 0, Path 1, Path 2)\\
- IMPORTANT: Agent name MUST be a quoted string, e.g., select\_path(0, ``\{target\_agent\}") NOT select\_path(0, \{target\_agent\})\\

EXAMPLE OUTPUT:\\
select\_path(0, ``\{target\_agent\}")\\
wait(2, 5, ``\{target\_agent\}")\\

Generate DSL:

\end{tcolorbox}

\subsubsection{Prompts for Real World Evaluation}

\begin{tcolorbox}[breakable]
You are a motion planning assistant that selects and modifies waypoint-based paths for a robot-human collaborative system. The robot and human are connected by a stick and must move together. You will receive waypoint information in text format, an image of the map showing the paths, and an instruction. \\

The system uses a dual-agent configuration: \\
- Robot (orange circle): The primary agent that is controlled. \\
- Human (cyan circle): Connected to robot by a stick, follows robot's movement. \\
- The robot's theta (rotation) determines the stick orientation. 0 degree means the horizontal positions with robot on the left. \\
Clockwise rotations increase the theta, and counter clockwise rotations decreases theta. \\

AVAILABLE DSL OPERATIONS: \\
1. select\_path(path\_index, agent): Select a path for the agent by index (0-based). Agent is "robot" or "human". \\
2. modify\_translation(step, dx, dy, agent): Move waypoint at step by (dx, dy) pixels. \\
3. modify\_rotation(step, degrees, agent): Rotate waypoint at step by degrees (+ clockwise, - counterclockwise). \\
4. insert\_waypoint(step, x, y, agent): Insert new waypoint after step at position (x, y). \\
5. wait(step, length, agent): Have agent wait at step for length additional steps. \\

The image shows: \\
- Colored lines with waypoint markers: Candidate paths (different colors = different homotopy classes) \\
- Brown blocks with names: Obstacles \\
- Grey blocks: Unreachable regions \\
- Orange/Cyan circles: Robot/Human start positions \\
- Green circles: Goal positions \\
- Numbers on waypoints indicate the waypoint index \\

COORDINATE SYSTEM: \\
- Origin (0,0) is top-left \\
- +x is right, +y is down \\
- Theta 0° = facing right, +90° = facing down \\

IMPORTANT RULES:
1. ALWAYS call select\_path() first for ``robot" to choose which path to use. \\
2. After select\_path, optionally modify the path to accommodate instructions with local movements. The original paths must be collision-free. If you have to make changes to accommodate the requirement in the instruction, try to keep a safe distance from any obstacles. \\
3. Only work on editing the robot path. \\
4. Use obstacle names for spatial reasoning. \\
5. Keep paths collision-free with obstacles and unreachable regions. \\

Please avoid modifying the first and last waypoint in a plan. They should be the start and the goal and always stay the same. Return ONLY the DSL code, no explanations or markdown formatting. \\

\end{tcolorbox} 

\subsection{Experiment Details}
In this section, we discuss the implementation details for the VirtualHome Human-Robot teaming, SimWorld Multi-Agent Cooperation and Real World Human-Robot Teaming for Joint Carrying experiments.

\subsubsection{VirtualHome Details}

In the VirtualHome experiments, the simulated human and the robot need to perform object rearrangement tasks together. We prompt Gemini 2.5 Pro to act as a simulated human: when the distance between the human and robot gets closer than a distance d (we use d = 5.0), the simulated human generates an instruction to the robot, given the current human and robot position and the human's remaining path. After one instruction generation, a cooldown period with n execution steps (we use n = 3) is applied before the next instruction. For the goal inference provided to \ours, we also prompt Gemini-2.5-Pro to select the most possible one given the full list of candidate receptacles, the human instruction, and the map state. The human and the robot are both assumed to have full map information and correct beliefs about the location of each object. We provided the exact prompts below:

\begin{tcolorbox}[breakable]
\textbf{Prompt for Instruction Generation} \\
You are a human working in a shared space with a robot. You need to give the robot a brief instruction to help coordinate your movements and avoid collisions.\\

CURRENT SITUATION:
- You (the human) are at position shown with a cyan circle\\
- The robot is at position shown with an orange circle\\
- Your planned path is shown as a blue line leading to your goal (green circle)\\
- You are currently {human\_goal\_desc}\\
- The robot's predicted path is also shown in red\\

The image shows:
- Brown blocks with names: Obstacles\\
- Grey blocks: Unreachable regions\\
- Blue line: Your remaining path\\
- Cyan circle: Your current position\\
- Orange circle: Robot's current position\\
- Green circles: Goals\\
- Red line: Robot's predicted path\\

Based on this situation, generate a SHORT instruction (1-2 sentences) for the robot to help coordinate movements. Avoid referring to things like 'your path' or 'my path' as robot do not know about them. For your instruction, avoid indicating that you will change the behavior. The robot should always change the behavior if in a conflict. If the robots stands near your future path, ask it to wait until you pass through (closer than 15 pixels). If the robot is standing in your way (path to your next waypoint) or really close to your future trajectory, ask it to move out of the way but do not refer to things like "path". If the robot's path seems to have no intersection with you, just instruct it to do it's thing. Make it clear for robot what you are trying to do and what you are hoping for them.\\

Return ONLY the instruction text, nothing else. \\

\end{tcolorbox} 

\begin{tcolorbox}[breakable]
\textbf{Prompt for Goal inference} \\
You are inferring the goal of an agent acting in some environment. Your task is to analyze an agent's trajectory and determine which goal location they are most likely heading towards. You will see an image, a human instruction, and a list of candidate goal locations. \\

The image shows:\\
- A map with obstacles (brown blocks with labels)\\
- The agent's past trajectory (a colored path/line showing where they have been walking)\\
- Various locations marked on the map\\

Based on the trajectory's direction, curvature, and progression, determine which of the following candidate goals the agent is most likely trying to reach:\\

CANDIDATE GOALS:\\
\{goals\_formatted\}\\

HUMAN INSTRUCTION:\\
\{instruction\}\\

Analyze the trajectory carefully:\\
1. Look at the overall direction of movement\\
2. Consider the most recent portion of the trajectory (where they are heading now)\\
3. Consider which goal location aligns best with the trajectory's direction\\

IMPORTANT: You must respond with ONLY the name of the selected goal from the list above. Do not include any explanation, reasoning, or additional text. Just output the goal name exactly as it appears in the list.\\

\end{tcolorbox} 

\subsubsection{SimWorld Details}
We consider two experimental settings for the multi-robot path coordination experiments: setups with two robots and with three robots. In the two-robot setup, we evaluate the models’ ability to navigate safely to the goal when starting from challenging configurations. In the three-robot setup, we randomize the starting locations of the robot cars. Whenever two cars come within 700 units of each other in Simworld, a negotiation is triggered. During negotiation, the robots take turns generating an utterance using Gemini 2.5 Pro. After each utterance is generated, a VLM provides an instruction to each robot to simulate input to \ours. After each negotiation, a cooldown period is applied for n execution steps, where we use n = 3, during which no new negotiation can be triggered. We use Gemini 2.5 Pro for goal inference and instruction generation. The robots are assumed to have full visibility of the map and the locations of all objects. We provide the exact prompts for goal inference, negotiation, and instruction generation below.

\begin{tcolorbox}[breakable]
\textbf{Prompt for Goal Inference}\\
You are analyzing agent trajectories in a parking lot environment to determine their goal destination.\\

ENVIRONMENT CONTEXT:\\
- This is a parking lot with numbered parking/charging spots\\
- Spots are labeled: 1, 2, 3, 4, 5, 6, 7, 8, 9, A, B, C\\
- Spot 8 is the EXIT (agents leaving the parking lot)\\
- All other spots (1-7, 9, A-C) are parking or charging spots\\

AGENT BEHAVIOR:\\
- Agents can move from one parking spot to another\\
- Agents can move from parking spots to charging spots or vice versa\\
- Agents can move to the exit (spot 8) to leave\\
- There is NO bias toward any particular type of destination\\
- Analyze only the trajectory direction and pattern\\

INSTRUCTION CONTEXT:\\
The other agent has given the following instruction to the robot to help coordinate movements.\\
other instruction:\\

This instruction may contain hints about where the other agent is heading. For example:\\
- ``I'm heading to the exit" suggests spot 8\\
- ``Let me get to my parking spot" suggests one of the parking spots\\
- ``I need to charge" suggests one of the charging spots\\
- References to direction (left, right, straight) may indicate the relative location of the goal\\
- If the instruction doesn't explicitly mention a destination, use it as context alongside the visual trajectory\\

YOUR TASK:\\
Look at the image and find the agent labeled \{agent\_name\}. This agent will have:\\
- A colored trajectory line showing their path\\
- A GREEN CIRCLE WITH SMALLER INNER CIRCLE at the START of the trajectory (where they began)\\
- A CYAN CIRCLE at the END of the trajectory (where they are now / their current position)\\
- A label with the text \{agent\_name\} (current) next to their current position\\

IMPORTANT - UNDERSTANDING THE TRAJECTORY:\\
- The SMALL GREEN CIRCLE shows where the agent STARTED\\
- The trajectory line shows the path they have taken FROM the start\\
- The CYAN CIRCLE shows where they are NOW (current position)\\
- The agent is moving AWAY FROM the small green circle (start) TOWARD their goal\\
- You need to determine where they are heading based on the direction of the trajectory\\

CRITICAL - SHORT OR MISSING TRAJECTORY:\\
- If the trajectory is very short or barely visible, it means the agent has just started moving and hasn't traveled far yet\\
- In this case, rely more heavily on the instruction to infer the goal\\
- Goals in the SAME ROW as the starting position are very unlikely - agents typically cross the parking lot\\
- The EXIT (spot 8) is a common destination when trajectory direction is unclear\\
- Do NOT assume the agent is heading to a nearby spot just because the trajectory is short\\

ANALYSIS GUIDELINES:\\
- FIRST, check if the instruction explicitly or implicitly mentions a destination\\
- SECOND, analyze the visual trajectory direction and curvature\\
- COMBINE both sources of information to make your best inference\\
- If the instruction strongly suggests a destination, weight it heavily\\
- If the instruction is ambiguous, rely more on the visual trajectory\\
- It is unlikely that the agent is heading toward a goal that shares the same row or column as the start\\
- When the trajectory is long enough, pay special attention to the most recent portion (near the cyan circle/current position)\\

Analyze THIS agent's trajectory carefully:\\
1. Consider what the human instruction reveals about their destination\\
2. Identify the trajectory belonging to \{agent\_name\} (look for the label)\\
3. Find the SMALL GREEN CIRCLE - this is the starting position\\
4. Follow the trajectory line from the start to see the direction of movement\\
5. Combine instruction context with visual analysis\\
6. Determine which numbered/lettered spot (1-9, A-C) the trajectory is pointing toward\\

CANDIDATE GOALS: 1, 2, 3, 4, 5, 6, 7, 8, 9, A, B, C\\

CRITICAL: You must respond with ONLY ONE of these goal labels: 1, 2, 3, 4, 5, 6, 7, 8, 9, A, B, or C\\

Do not include any explanation, reasoning, or additional text. Just output the single goal label.
\end{tcolorbox}

\begin{tcolorbox}[breakable]
\textbf{Prompt for Negotiation Utterance Generation}\\
You are \{agent\_name\}, a robot navigating a parking lot. You are in a potential collision situation with other robots and need to communicate your intentions.\\

WHAT YOU CAN SEE IN THE IMAGE:
- Your current position (labeled ``\{agent\_name\} (me)")\\
- Your goal marker (labeled ``My Goal")\\
- Your planned path (colored line from your position toward your goal)\\
- Other robots' current positions and orientations (labeled with their names)\\
- You can NOT see other robots' goals or planned paths\\

OTHER ROBOTS IN THIS NEGOTIATION: \{', '.join(other\_agents)\}\\

\{prev\_text\}\\

YOUR TASK:\\
Generate a short utterance (1-2 sentences) that:\\
1. Declares your intent using RELATIVE DIRECTIONS (e.g., ``heading toward the upper area", ``going left", ``moving toward the far end")\\
2. Does NOT mention specific spot numbers or coordinates\\
3. Suggests a coordination strategy if appropriate (e.g., ``I can wait", ``please let me pass first", ``I'll go around")\\
4. Considers the relative positions of other robots you can see\\
5. Suggest movements for other agents as needed (i.e. move backwards a bit)\\

IMPORTANT:\\
- Use relative terms: upper, lower, left, right, center, near, far, diagonal, etc.\\
- Be concise - this is a quick communication, not a detailed explanation\\
- You're a robot, speak naturally but briefly\\

Return ONLY the utterance text, nothing else.
\end{tcolorbox}

\begin{tcolorbox}[breakable]
\textbf{Instruction Generation Prompt}\\
You are a traffic coordinator for autonomous robots in a parking lot. Multiple robots are close to each other and risk collision. They have each stated their intentions. Your job is to give each robot a clear instruction to avoid a collision.\\

SITUATION:\\
\{len(negotiation\_group)\} robots are in close proximity and need coordination.\\

CURRENT POSITIONS (world coordinates):\\
\{positions\_section\}\\

ROBOT UTTERANCES (their stated intentions):\\
\{utterances\_text\}\\

IMAGES PROVIDED:\\
You are shown a 2x2 grid of images. Each image shows ONE robot's perspective:\\
- That robot's candidate paths (solid colored lines labeled Path 1, Path 2, Path 3)\\
- Other robots' predicted paths (dashed lines)\\
- All robot positions and orientations\\

Use these visualizations to understand the spatial relationships and make coordination decisions.\\

YOUR TASK:\\
Provide a specific instruction for EACH robot to help them avoid collision.\\
Instructions should:\\
1. Be actionable (wait, proceed, take alternate route, slow down, etc.)\\
2. Consider each robot's stated intention and available paths\\
3. Be fair - don't always make the same robot wait\\
4. Use relative directions (left, right, forward, around) not absolute coordinates\\
5. Be concise (1-2 sentences per robot)\\
6. Reference specific path numbers when appropriate (e.g., ``take Path 2")\\
7. Tell the robots to move as needed (i.e. move forward/back)\\

IMPORTANT CONSTRAINTS:\\
- Each robot can only modify its OWN path\\
- Instructions should help avoid collisions while respecting intentions\\
- Consider the spatial layout shown in the images\\
- Make sure to stay true to the negotiation utterances given by the robots\\

FORMAT YOUR RESPONSE EXACTLY AS:\\
{negotiation\_group[0]}: [instruction for {negotiation\_group[0]}]\\
{negotiation\_group[1] if len(negotiation\_group) $>$ 1 else `'}: [instruction for {negotiation\_group[1] if len(negotiation\_group) $>$ 1 else `'}]\\
...and so on for each robot\\

ONLY output the instructions in the format above, nothing else.
\end{tcolorbox}

\subsubsection{Real World Experiment Details}

We use the Stretch 3 robot to cooperate with participants in all human experiments. All the experiments are run on 4 pre-designed maps. The human and the robot need to carry a PVsC pipe together from the start to the goal on two sides of an open space, getting through a region cluttered with obstacles. Table \ref{tab:comp} shows the comparison between our experimental setup and another work \cite{yang2025implicitcommunicationhumanrobotcollaborative} with human-robot joint carrying experiments. The task for that work is to have the human-robot pair get through a single obstacle in the center of the space. Comparing with the previous works, our setup is significantly more challenging, with more diverse and challenging obstacle patterns in maps. 

Before the experiment, we construct a 2D map for each map setup. We use the navigation and map construction module of stretch robot. With teleoperation control, we have the robot scan the region with lidar and build an occupancy grid, from which we construct the map with the name and the location of each obstacle. During the experiment, the participants instructed the robot with an attached USB microphone. The recording was triggered with the wake word ``robot" and recorded for the next 8 seconds. For each new instruction, the robot stops at the position, accesses the current location with builtin localization module, and replans with different methods to get a sequence of waypoints (x, y, $\theta$) to navigate to. The robot then navigates to each waypoint in a straight line. For the navigation, we impose a safety margin of 10cm. The experiment failed whenever the robot got closer to an obstacle than that threshold during the execution. Both the participant and the robot know the exact location of the destination.

During the experiment, the participant consecutively performed 4 tasks with different models. Between each trial, we asked the participant to rate the models by answering the following questions on a scale of 1 to 7: 1. The robot understood my instructions 2. The robot's path was efficient and reasonable 3. The robot always chose safe paths. After all four trials, we ask the participant to rank the four models in the order of the experiments. 

\begin{table}[t!]
\centering
\begin{tabular}{ccc}
\hline
Metrics & Ours & IC-MPC \\
\hline
Object length & 0.85m & 0.914m \\
Area size & 36$m^2$ & $15.68m^2$ \\
Map number & 4 & 1 \\
Obstacle number & 4-7 & 1 \\
\hline
\end{tabular}
\caption{Comparing our real experiments setup with IC-MPC}
\label{tab:comp}
\end{table}

\subsection{Training Details}
We finetuned the Qwen-2.5-7B model as the motion program synthesizer. We use all synthetic data for our training data. Each entry of training data contains a map marking the human path and a set of possible paths returned by the planner, an instruction,
and the label code for the correct modification. To generate our maps, we uniformly sample obstacle locations at random on a 1000 * 1000 image, with the width and height of each obstacle between 20 and 50 pixels. In our data, we have three levels of object cluttering and object densities within the map. In total, we generate 5000 maps. To train specific behaviors with narrow space, we also add 1000 maps with specifically designed general structures like narrow corridors or crossroads. For each of the maps, we sample 5 random start and goal locations for the human and the robot and use the planner to plan the predicted human path \& the set of paths for the robot to show in the image. To generate specific data, we define the following set of behaviors: movements (left, right, forward, backward, from either the human or the robot perspective), rotation (clockwise or counterclockwise), path selection (different ways of using landmarks), obstacles (stay away or closer to certain objects), wait (when paths cross) and backout (when robots stands in the way of human). For each of these coordination behaviors, we generate the templated instructions and GT path edit code with our DSL. In total, we have 40k training data. The training runs on 4 A100 GPUS for 36 hours.

\subsection{Pilot Study}
In order to design the DSL for \ours, we conducted a pilot study to determine what types of instructions humans might give in the real world. We recruited 10 participants from a college campus and paired them up. For each pair, they first perform a human-human joint carrying task with a long PVC pipe, where they navigated a challenging map with tight gaps to evaluate the types of commands people give each other during joint lifting. The participants then performed a human-robot trial, where one participant carried one end of the stick, and the other participant controlled the robot with a controller. During the session, we observed that participants gave instructions that included both global directives like ``go between the sofa and the wall", and local edits like ``Stay away from the door", ``move forward a bit into the room, then rotate right". These instructions required both high-level and low-level adjustments, which motivated the design of CaPE. We designed the DSL to be simple yet comprehensive, making sure that the path can be modified to follow any participant's instructions with the combination of these operations. The pilot study was approved by an institutional review board.

\end{document}